\documentclass{article}

\PassOptionsToPackage{numbers, compress}{natbib}
    \usepackage[preprint]{neurips_2023}

\usepackage[utf8]{inputenc} % allow utf-8 input
\usepackage[T1]{fontenc}    % use 8-bit T1 fonts
\usepackage[pagebackref,breaklinks,colorlinks]{hyperref}
\usepackage{url}            % simple URL typesetting
\usepackage{booktabs}       % professional-quality tables
\usepackage{amsfonts}       % blackboard math symbols
\usepackage{nicefrac}       % compact symbols for 1/2, etc.
\usepackage{microtype}      % microtypography
\usepackage{xcolor}         % colors

\usepackage{amsmath}
\usepackage{amsfonts}
\usepackage{amssymb}
\usepackage{wrapfig}
\usepackage{subcaption}
\usepackage{multirow}
 \usepackage{mathtools} 

\usepackage{verbatim}

\usepackage{anyfontsize}

\usepackage{microtype}
\usepackage{graphicx}
\usepackage{booktabs} % for professional tables
\usepackage{multirow}
\usepackage{amsmath,amssymb}
\usepackage{booktabs}
\usepackage{caption,subcaption}

\usepackage{xcolor}
\definecolor{mygreen}{HTML}{3cb44b}
\definecolor{skyblue}{HTML}{beffff}
\definecolor{lightgreen}{HTML}{90ee90}

\usepackage{color, colortbl}

\definecolor{emerald}{rgb}{0.31, 0.78, 0.37}

\usepackage{tcolorbox}
\usepackage{enumitem}
\setitemize{itemsep=10pt,topsep=0pt,parsep=0pt,partopsep=0pt}
\usepackage{colortbl}

\usepackage{xcolor}
\definecolor{mygreen}{HTML}{3cb44b}
\colorlet{myyellow}{green!10!orange!90!}
\makeatletter

\usepackage{tikz}
\usetikzlibrary{arrows,shapes,automata,backgrounds,fit,petri}
\usepackage{adjustbox}

\newcommand{\RN}[1]{%
	\textup{\lowercase\expandafter{\it \romannumeral#1}}%
}
\usepackage{tabu}
\newcommand{\eg}[0]{\emph{e.g., }}

\newcommand{\beq}{\vspace{0mm}\begin{equation}}
\newcommand{\eeq}{\vspace{0mm}\end{equation}}
\newcommand{\beqs}{\vspace{0mm}\begin{eqnarray}}
\newcommand{\eeqs}{\vspace{0mm}\end{eqnarray}}
\newcommand{\barr}{\begin{array}}
\newcommand{\earr}{\end{array}}

\usepackage{color, colortbl}
\definecolor{Gray}{gray}{0.93}

\usepackage{lipsum}

\usepackage{pifont}% http://ctan.org/pkg/pifont
\newcommand{\cmark}{\ding{51}}%
\newcommand{\xmark}{\ding{55}}%

\usepackage{makecell}

\usepackage{xcolor,amsmath}
\usepackage[linesnumbered,ruled,vlined]{algorithm2e}
\DontPrintSemicolon

\usepackage{xcolor}
\definecolor{mygreen}{HTML}{3cb44b}

\SetKwComment{Comment}{\color{green!50!black}\# }{}

\SetKwProg{Function}{def}{:}{}

\SetKwProg{For}{for}{:}{}
\SetKwProg{If}{if}{:}{}

\usepackage{xcolor}

\newcommand{\shortname}{LLaVA}

\title{An Empirical Study of Scaling Instruction-Tuned\\ Large Multimodal Models }

\author{Yadong Lu$^{*1}$, Chunyuan Li$^{*2}$, Haotian Liu$^{3}$, Jianwei Yang$^{2}$, Jianfeng Gao$^{2}$,  Yelong Shen$^{1}$ \\
$^1$Microsoft Azure AI~~~ $^2$Microsoft Research~~~ $^3$University of Wisconsin--Madison  \\
 \texttt{} }

\begin{document}

\def\thefootnote{*}\footnotetext{These authors contributed equally to this work}\def\thefootnote{\arabic{footnote}}

\maketitle

\begin{abstract}

Visual instruction tuning has recently shown encouraging progress with open-source large multimodal models (LMM) such as LLaVA and MiniGPT-4. However, most existing studies of open-source LMM are performed using models with 13B parameters or smaller. In this paper we present an empirical study of scaling LLaVA up to 33B and 65B/70B, and share our findings from our explorations in image resolution, data mixing and parameter-efficient training methods such as LoRA/QLoRA. These are evaluated by their impact on the multi-modal and language capabilities when completing real-world tasks in the wild. 
We find that scaling LMM consistently enhances model performance and improves language capabilities, and performance of LoRA/QLoRA tuning of LMM are comparable to the performance of full-model fine-tuning. Additionally, the study highlights the importance of higher image resolutions and mixing multimodal-language data to improve LMM performance, and visual instruction tuning can sometimes improve LMM's pure language capability. We hope this study makes state-of-the-art LMM research at a larger scale more accessible, thus helping establish stronger baselines for future research. Code and checkpoints will be made public.

% In this note, we provide a concise overview of our experimental observations and insights gained from the process of upscaling Vision-Language models. Our experiments centered on the instruct finetuning of VL models with 33B and 65B LLMs, followed by an evaluation of the these models on three distinct benchmarks: LLaVA Bench, MMBench, and MM-Vet. The key findings can be summarized as follows:
% 1) The act of scaling the language models exhibits a consistent positive impact on the overall performance of vision-language models across the various evaluation benchmarks. However, it's important to note that the optimal training schedules differ across different model sizes.
% 2) Our observations indicate the crucial role of instruction finetuning data in achieving high performance. 
% Intriguingly, we found that the 33B model, when trained with a larger volume of data, outperforms the 65B model.
% 2) We conducted on a qualitative analysis comparing responses from models of different sizes, unveiling both the enhanced capabilities enabled by scaling and the domains in which scaling offers no discernible benefits. 3) we conducted a study of data mixing to see its impact on the performance.

\end{abstract}

\section{Introduction}
% The intersection of vision and language has yielded remarkable advancements in artificial intelligence, paving the way for models capable of comprehending and generating textual and visual information. This synergy has been further propelled by the advent of large-scale language models (LLMs), which have proven instrumental in capturing intricate linguistic patterns and contextual nuances. As the field continues to evolve, there is an increasing emphasis on upscaling these models to harness their potential for more complex tasks and nuanced understanding.

% In this context, this technical report presents a comprehensive exploration of the implications and outcomes of scaling Vision-Language models. Our investigation involves the finetuning for end-to-end integration of vision encoder alongside LLMs of varying magnitudes, particularly focusing on models with 33B and 65B parameters. Through a meticulous series of experiments and evaluations, we seek to unravel the nuances of scaling and its effects on performance across multiple evaluation benchmarks.

Recent studies on large multimodal models (LMM)~\citep{li2023large,li2023multimodal} have been focused on 
%are converging on a central concept known as 
the methods of {\it visual instruction tuning}~\citep{llava}. The
results are promising: \eg the open-source project Large Language and Vision Assistant (\shortname{}) shows that training a 7B large language model (LLM) with multimodal instruction-following data for 3 hours on 8 A-100 GPUs leads to a LMM with strong visual understanding and reasoning capabilities in the wild: reproducing some of the most appealing examples of the proprietary OpenAI multimodal GPT-4 model~\citep{gpt4}. 
A similar idea is explored in its co-current work MiniGPT-4~\citep{zhu2023minigpt}. It has rapidly become a prominent research topic, spurring the development of numerous new models, benchmarks, and applications~\citep{li2023multimodal}. However, the high compute cost has led most existing studies to utilize 7B and 13B LLMs. Thus, the impact of significantly scaling up the model size to \eg 33B and 65B remains unexplored.

This study aims to fill this gap by empirically investigating language models of larger sizes for LMM, sharing insights of our scaling experiments and establishing stronger baselines using larger-scale LLaVA for future research. Specifically, we explore the impact of larger model sizes, model tuning and data mixing methods on model performance, and present our findings and recommendations. The scaling recipe leads to new state-of-the-art (SoTA) performance on LLaVA-Bench~\citep{llava} and MM-VET~\cite{yu2023mm}. We hope that our findings and larger LLaVA checkpoints would provide a reference for future research on visual instruction tuning.

\section{Experiment Setup}
\label{sec:headings}
% 1. models  2. training schedule, 3. data

\paragraph{Model Checkpoints.} To study the impact of scaling up LLM on multimmodal capabilities, we increase the language model size to 33B and 65B~\citep{touvron2023llama}, in addition to the 7B and 13B models used for existing LMM.

\begin{itemize}[leftmargin=7.5mm]
\setlength{\itemsep}{2pt}
\item {\bf \shortname{}-${\text{33B}}$}~~~We employ the open source Vicuna-33B checkpoint~\footnote{\url{https://huggingface.co/lmsys/vicuna-33b-v1.3}}~\citep{vicuna} to preform the two-stage training. The training data is around 125K conversations collected from \url{ShareGPT.com}.

\item {\bf \shortname{}-${\text{65B}}$}~~~Due to a lack of public 65B Vicuna checkpoint, we conduct our own training of the Vicuna-65B model, utilizing ShareGPT data that we have independently processed. This data contains 159M tokens used during training. As a comparison, the reported number of tokens used in training Vicuna 33B is 370M ~\footnote{\url{https://github.com/lm-sys/FastChat/blob/main/docs/vicuna_weights_version.md}}. 

%  in total 156K instruct tuning format samples from ShareGPT website, which corresponds to

% \yd{I updated the stats here} \chunyl{The stats of our own ShareGPT data and vicuna shareGPT data are not in the same units. It is import to clearly report the difference.}
\end{itemize}

Once the instruction-tuned LLM is given, we follow \cite{llava} to perform the two-stage LLaVA lightning training:  
$(i)$ {\it Stage 1: Pre-training for Feature Alignment.} The linear projection layer is trained, which maps the visual feature (the features before the last layer of the pre-trained image encoder) to  word embedding space of LLM. More specifcally, the projection dimension is 1024$\rightarrow$6656 for the 33B model and 1024$\rightarrow$8192 for the 65B model, respectively. In this stage, we use the concept-balanced subset of LAION-CC-SBU data with 558K samples. 
$(ii)$ {\it Stage 2: Visual Instruction Tuning.} We use the LLaVA-80K multimodal instruct dataset for the fine-tuning stage. 
Various training schedules are explored to enable the model to follow the diverse instructions to complete tasks in the wild, as to be detailed below.

\paragraph{Tuning Methods.} We explore both the trainable modules and training data mixing for efficient and effective visual instruct tuning of large models.

\begin{itemize}[leftmargin=7.5mm]
\setlength{\itemsep}{2pt}
\item {\bf Trainable modules.}~~~In addition to tuning the linear projection layer, two schemes are considered to tune the LLM: $(i)$ Full-model fine-tuning of LLM and $(ii)$ Parameter-efficient training methods. For the latter, LoRA~\citep{hu2021lora} and QLoRA~\citep{dettmers2023qlora} are employed to allow us to tune large models with limited compute resource. This aims to gain an in-depth understanding of the trade-off between the training cost and model performance.

\item {\bf Data mixing.}~~~ Typically only the multimodal instruction data is used in Stage-2. We further consider mixing the language-only instruct data ShareGPT with the LLaVA-80K multimodal instruction data to gain an in-depth understanding of the trade-off between models' language and multimodal capabilities. 
% \chunyl{Report which visual instruct data is employed}
\end{itemize}

\paragraph{Hyper-parameters.}
In the training process of both stages, we utilize the DeepSpeed library \footnote{\url{https://github.com/microsoft/DeepSpeed}} and employ the ZeRO3 optimizer, except for QLoRA runs we use ZeRO2. We use a maximum sequence length of 2048. 
For Stage 1, we train both the 33B and 65B models with a learning rate of $1\!\times\!10^{-4}$ with
no weight decay, and a learning rate with linear decay and linear warmup for 3\% of training steps in total. 
For Stage 2,
we use a learning rate of $2\!\times\!10^{-5}$ in full fine-tuning to train 1 epoch for all the models in full finetuning, and a learning rate of $1\!\times\!10^{-4}$ for the LoRA/QLoRA runs. We conducted a set of hyperparameter search and for LoRA runs, and found larger LoRA alpha or equivalently larger learning rate was crucial to get the best performance. Specifically, we use LoRA alpha equals 2 times the LoRA rank, and a learning rate of $1\!\times\!10^{-4}$, which works the best for all the models. For full fine-tuning, we use a total batch size of 512 on 4 A100 nodes, where each of these nodes is equipped with 8 A100-80G GPUs. For LoRA/QLoRA runs, we use a total batchsize of 64 on 1 A100 node for 33B model and 2 nodes for 65B model.

% We 
% We find the training schedule is essential for our model size scaling experiments. In particular, we find increased resolution of vision model helps the performance and batchsize improves the results. 

% \chunyl{Empirical insights: LLM Model scaling with full-model tuning, LoRA and QLoRA: performance and cost.}

% \chunyl{Increase of image resolution benefits visual recognition, and increase of LLM model size benefits knowledge/language ability.}

\section{Results}
% 1. benchmark results 2. ablation of training vision encoder or not 3. effect of 224 vs 336 4. training schedule 

% ckpts
% vanilla fft
% 33B+65B: 224+336 (4 models)
% 65B 150k data 336 

% lora 
% rank 64, no mixing: 33B+65B (2 models)
% rank 128, rank 512: 65B (2 models)

% data mixing
% 33B+65B: maxlen 1024 vs 2048 (4 models)

% Our results can be categorized into 3 sections: 1) We train 33B model and 65B model with resolution 224x224 and 336x336 ViT image encoder (4 models), 2) we train 33B and 65B model with LoRA rank 64, and further study the effect of increasing the LoRA rank to 128 and 512 for 65B model (4 models) and 3) we conduct data mixing study by training 33B and 65B model with max length 1024 and 2048 (4 models).

%We have trained both 33B and 65B models for 1 epoch with a learning rate of 2e-5, and batch size 512. \chunyl{sometimes we report 1K or 2K batch sizes, please specify in what settings we consider them}

We first compare our large checkpoints on two recent benchmarks which are specifically designed for LMM, then report our findings in the course of scaling up LLaVA models.

\begin{table}[t!]
    \centering
    \begin{tabular}{l|ccc|c}
        \toprule
Models & Reasoning&	Conversation & 	Detail& 	 	Overall \\ \midrule
Bard-0718 & 78.7 &	83.7 & 	69.7 & 	 	77.8 \\
Bing-Chat-0629 & 90.1& 	59.6& 	52.2& 		71.5 \\  \midrule
LLaVA-13B  (beam=1)	& 81.7	&  64.3	& 55.9	& 70.1\\
LLaVA-13B  (beam=5)	& 84.3	& 68.4	& 59.9	&  73.5 \\
\rowcolor{Gray}
LLaVA-33B  (beam=1)	& 82.9 & 70.2 & 62.6 & 73.9 \\
\rowcolor{Gray} 
LLaVA-33B  (beam=5)	& 83.5 & 72.6 & 61.9 & 74.8 \\
\rowcolor{Gray}
LLaVA-65B  (beam=1)	&  87.3 & 63.8 & 62.3 & 74.2 \\
\rowcolor{Gray}
LLaVA-65B  (beam=5)	& 88.7 	&  59.4	& 65.7 	& 74.4  \\ 
\bottomrule
    \end{tabular}
    \vspace{1mm}
    \caption{The performance comparison on LLaVA-Bench. Beam search sizes at 1 and 5 are reported. 
    }
    \label{tab:sota_on_llava_bench}
\end{table}

\begin{table}[htbp]
\centering
\resizebox{0.99\textwidth}{!}{%
\begin{tabular}{l|cccccc|c}
\toprule
Model & Rec & OCR & Knowledge & Generation & Spatial & Math & Total \\
\midrule
 \multicolumn{8}{l}{\it Results of various open-source LMM on reported in the MM-VET paper~\cite{yu2023mm}} \\
% Transformers Agent (GPT-4) [34] & 18.2 & 3.9 & 2.2 & 3.2 & 12.4 & 4.0 & 13.4±0.5 \\
LLaMA-Adapter v2-7B~\cite{gao2023llama} & 16.8 & 7.8 & 2.5 & 3.0 & 16.6 & 4.4 & 13.6±0.2 \\
OpenFlamingo-9B~\cite{alayrac2022flamingo,awadalla2023openflamingo} & 24.6 & 14.4 & 13.0 & 12.3 & 18.0 & 15.0 & 21.8±0.1 \\
MiniGPT-4-8B~\cite{zhu2023minigpt} & 27.4 & 15.0 & 12.8 & 13.9 & 20.3 & 7.7 & 22.1±0.1 \\
BLIP-2-12B~\cite{li2023blip} & 27.5 & 11.1 & 11.8 & 7.0 & 16.2 & 5.8 & 22.4±0.2 \\
LLaVA-7B~\cite{llava} & 28.0 & 17.1 & 16.3 & 18.9 & 21.2 & 11.5 & 23.8±0.6 \\
MiniGPT-4-14B~\cite{zhu2023minigpt} & 29.9 & 16.1 & 20.4 & 22.1 & 22.2 & 3.8 & 24.4±0.4 \\
Otter-9B~\cite{li2023otter} & 28.4 & 16.4 & 19.4 & 20.7 & 19.3 & 15.0 & 24.6±0.2 \\
InstructBLIP-14B~\cite{dai2023instructblip} & 30.8 & 16.0 & 9.8 & 9.0 & 21.1 & 10.5 & 25.6±0.3 \\
InstructBLIP-8B~\cite{dai2023instructblip} & 32.4 & 14.6 & 16.5 & 18.2 & 18.6 & 7.7 & 26.2±0.2 \\
LLaVA-13B~\cite{llava} & 30.9 & 20.1 & 23.5 & 26.4 & 24.3 & 7.7 & 26.4±0.1 \\
\textcolor{gray}{MM-ReAct-GPT-3.5}~\cite{yang2023mmreact} & \textcolor{gray}{24.2} & \textcolor{gray}{31.5} & \textcolor{gray}{21.5} & \textcolor{gray}{20.7} & \textcolor{gray}{32.3} & \textcolor{gray}{26.2} & \textcolor{gray}{27.9±0.1} \\
LLaVA-7B (LLaMA-2)~\cite{llava} & 32.9 & 20.1 & 19.0 & 20.1 & 25.7 & 5.2 & 28.1±0.4 \\
LLaVA-13B (V1.3, 336px)~\cite{llava} & 38.1 & 22.3 & 25.2 & 25.8 & 31.3 & 11.2 & 32.5±0.1 \\
LLaVA-13B (LLaMA-2)~\cite{llava} & 39.2 & 22.7 & 26.5 & 29.3 & 29.6 & 7.7 & 32.9±0.1 \\
\textcolor{gray}{MM-ReAct-GPT-4}~\cite{yang2023mmreact}  & \textcolor{gray}{33.1} & \textcolor{gray}{65.7} & \textcolor{gray}{29.0} & \textcolor{gray}{35.0} & \textcolor{gray}{56.8} & \textcolor{gray}{69.2} & \textcolor{gray}{44.6±0.2} \\
\midrule
 \multicolumn{8}{l}{\it Results with our own experiment runs } \\
\rowcolor{Gray}
LLaVA-13B (LLaMA-2)& 38.4 & 21.0 & 26.3 & 28.8 & 28.0 & 7.7 & 32.6±0.1  \\
\rowcolor{Gray}
\shortname{}-33B & 38.5 & 25.0 & 26.2 & 28.2 & 29.2 & 7.7 & 32.9±0.3  \\
\rowcolor{Gray}
\shortname{}-33B (Data Mixing) & 37.7 & 27.1 & 26.2 & 28.6 & 28.1 & 11.5 & 34.1±0.3   \\
% \rowcolor{Gray}
% \shortname{}-33B  (Mix, unfreeze vision) & 38.4 & 21.3 & 24.0 & 27.9 & 28.9 & 3.8 & 31.6  \\
\rowcolor{Gray}
\shortname{}-65B & 39.2 & 28.2 & 26.2 & 28.3 & 33.0 & 15.0 & 35.5±0.3  \\
\rowcolor{Gray}
\shortname{}-65B  (Data Mixing) & 41.8 & 27.9 & 30.4 & 32.3 & 30.5 & 7.3 & \textbf{36.4±0.2}  \\
% \rowcolor{Gray}
% \shortname{}-70B  (Mix) & 39.3 & 22.9 & 24.5 & 25.9 & 32.0 & 11.5 & 32.9 \\
\bottomrule
\end{tabular}
}
\vspace{1mm}
\caption{Performance of various open-source LMM on MM-VET. Note that MM-ReAct is not an single multimodal model, it is a system built on chaining visual tools via GPT-3.5 or GPT-4, which we append as a reference. Our experiment run on LLaVA-13B (LLaMA-2) yields very similar score with the same checkpoint reported in MM-VET paper, indicating that our evaluation pipelines are consistent.}
\label{tab:mmvet}
\end{table}

\subsection{Comparisons on Benchmarks}

\paragraph{LLaVA-Bench.}  LLaVA-Bench (In-the-Wild)\footnote{\url{https://github.com/haotian-liu/LLaVA/blob/main/docs/LLaVA_Bench.md}}~\cite{llava} is a diverse evaluation dataset consisting of 24 images with 60 questions in total, including indoor and outdoor scenes, memes, paintings, sketches. Each image is paired with a manually-curated, detailed description and a set of properly-selected questions related to open-ended visual chat scenarios. 
Each questions belongs to one of three types of tasks: conversations that contain simple visual recognition \& QA questions, detailed descriptions that characterize the image with a long paragraph, and a complex reasoning task that focuses on deducing implications from an image. 
Language GPT-4 ($\texttt{gpt4-0314}$) is used to score to the generated answers. The relative scores between the model output and gold response are reported.
We compare LLaVA against the commercial visual chat systems including Microsoft BingChat\footnote{\url{https://www.bing.com/chat}} and Google Bard\footnote{\url{https://bard.google.com/}} on LLaVA-Bench~\cite{llava}. 

The results are presented in Table~\ref{tab:sota_on_llava_bench}.
The 33B and 65B checkpoints outperform the 13B LLaVA model and Bing Chat. Despite the fact that LLaVA-Bench is small (thus the comparison might not be statistically significant), the results are encouraging: compared to large LMM, small open-sourced LMM are far more cost-effective to be deployed in real-world applications. %for more efficient inference. 
With negligible increase of inference latency, we can significantly improve the performance for all model sizes by increasing the beam search size from 1 to 5. 
Our results show that larger LLaVA models generally exhibit better performance in tasks involving complex reasoning and generating detailed descriptions, which requires strong language competencies from larger LLM. In addition, larger LLaVA models obtain comparable results to BingChat in multi-turn, multi-modal conversation tasks that require strong image understanding capability.

\paragraph{MM-VET.} 
MM-VET~\cite{yu2023mm} is designed based on the assumption that the intriguing capability of solving complicated tasks is often achieved by a generalist LMM which is able to integrate a varity of vision-language (VL) capabilities. MM-Vet contains 200 images and 218 questions (samples), aiming to evaluate6 core VL capabilities (recognition, OCR, knowledge, language generation, spatial awareness, and math) and their combinations. 
%examining the 16 integrations of interest derived from the capability combination. 
For evaluation, an LLM-based evaluator ($\texttt{gpt4-0613}$) is used to score open-ended outputs of different forms.
%The evaluator can score answers of different forms, resulting in a unified scoring metric.
%
In Table~\ref{tab:mmvet}, we report the results on MM-VET. 
The performance is consistently improved from 13B to 33B and 65B. The largest LLaVA model improves SoTA performance among the end-to-end open-source LMM. The most significant improvements are observed when evaluating the capabilities of knowledge and generation, followed by recognition and OCR. The performance on spatial and math remains comparable. The result reveals that the improved LLM capability is instrumental in storing more knowledge in the weights and leading to a stronger language responding capability. % thereby benefiting the knowledge and language categories. 

% \paragraph{VisIT-Bench.} VisIT-Bench~\cite{bitton2023visitbench}

\subsection{Scaling up \shortname{}}
The experiments are conducted to answer three research questions.

\paragraph{\textcircled{\raisebox{-0.9pt}{1}} Which scaling factor matters?} 
We study the relative contribution of three scaling-up factors to the performance improvement of \shortname{}. The results are summarized in Table~\ref{tab:scaling_on_llava_bench} (a). 
\begin{itemize}[leftmargin=7.5mm]
\setlength{\itemsep}{2pt}
\item {\bf Model size.}~~~ Increasing the model size consistently improves the overall performance. We conjecture that larger data size is essential to train a larger model.  For example, if we only train on LLaVA-80K data, we see smaller gain when model size becomes larger.
% i.e., increasing the model size gradually from 7B to 13B, 33B, and 65B yields gains of approximately 4, 2, and 1 points, respectively. 
\item {\bf Image resolution.}~~~By fixing the CLIP ViT image encoder, we compare the variants that are pre-trained to take image resolution $224\times224$ and $336\times336$, and find that the higher resolution consistently yields 2-3 points improvement across all four LLM sizes.
\item {\bf Data mixing.}~~~Larger models tend to have higher capability of fitting the instruction data. By mixing the language-only instruction data (ShareGPT) with LLaVA-80K, we can improve model performance by 2 points, compared to training on multimodal instruction data only.
\end{itemize}

In Table~\ref{tab:scaling_on_llava_bench} (b), we present our result on MM-Bench~\citep{liu2023mmbench}, which contains a set of 2,974 questions, which evaluate models' reasoning skills of six categories. The combination of the three factors improve the baseline LLaVA 7B model, reported in~\cite{liu2023mmbench}.

\begin{table}[t!]
    \centering
    \begin{subtable}{0.99\textwidth} 
    \centering 
    \resizebox{0.60\textwidth}{!}{%
    \begin{tabular}{lccccc}
        \toprule
      Image Size & Data Mixing  &  7B  &  13B &  33B  &  65B  \\
       \midrule
        224$\times$224 & \xmark   & 63.6 & 67.1 & 69.3 & 70.3  \\
        336$\times$336 & \xmark   & 65.9 & 70.1 & 72.0 & 72.3  \\
        336$\times$336&  \cmark  & -- & -- & 73.9 & 74.2 \\
       \bottomrule
    \end{tabular}
    }
    \vspace{0mm}
\caption{Performance scores on LLaVA-Bench.}
\end{subtable}

\begin{subtable}{0.99\textwidth} 
\centering 
\resizebox{0.99\textwidth}{!}{%
\begin{tabular}{lrcccccccc}
\toprule
Checkpoint & Image Size & Data Mixing & Overall & LR & AR & RR & FP-S & FP-C & CP \\
\midrule
\shortname{}-7B & 224$\times$224 & \xmark& 36.2 & 15.9 & 53.6 & 28.6 & 41.8 & 20.0 & 40.4 \\

\shortname{}-33B & 336$\times$336& \cmark & 55.7 & 23.3  & 74.0 & 46.0  & 51.5 & 50.4 & 67.2 \\
% \textbf{Ours-33B mix} & 33.2B & Vicuna-33B & CLIP ViT-L/14 & 55.6 & 24.4  & 72.7 & 48.4  & 53.5 & 47.2 & 66.2 \\

\shortname{}-65B & 336$\times$336 & \cmark & 56.0 & 24.4  & 72.3 & 49.3  & 50.5 & 51.2 & 68.1 \\
\bottomrule
\end{tabular} }
\vspace{0mm}
\caption{Performance scores on MM-Bench. The skills to evaluate include logic reasoning (LR), attribute reasoning (AR), relation reasoning (RR), fine-grained single-instance perception (FP-S), fine-grained cross-instance perception (FP-C), and coarse perception (CP).}
\end{subtable}

    \caption{The performance to scale up model size, image resolution and data mixing.  
    % \chunyl{Please report the training epoch, learning rate and number of GPUs so that we could ensure all checkpoints are fairly compared.} 
    }
    \label{tab:scaling_on_llava_bench}
\end{table}

\paragraph{\textcircled{\raisebox{-0.9pt}{2}}  When should the parameter-efficient training method be considered?}
As model size increases, it becomes necessary to consider using tuning methods that are more efficient than full-model fine-tuning. LoRA and QLoRA are well-known parameter-efficient tuning methods. 
%We compare them in Table~\ref{tab:tuning_methods}.
As shown in Table~\ref{tab:tuning_methods}, we report compute cost using {\it GPU hours per node}, because the unit can be equivalent to the price \$13.63/hour (ND A100 v4 series) on Azure~\footnote{\url{https://azure.microsoft.com/en-us/pricing/details/machine-learning/}}. The total cost can be estimated by multiplying the \#hours and \#epochs.

In Table~\ref{tab:tuning_methods}(a), we train both the 33B and 65B model with LoRA rank 8 and 64 for 1 epoch on the LLaVA-80K instruction-tuning dataset. For models with 33B parameters and above, as we increase the LoRA rank values, we notice an increase in both performance and cost until full-model tuning reaches its maximum performance for a specific model size. In the case of the 13B model, we find that a rank of 64 can deliver comparable performance to full-model tuning. The cost is more related to the total number of parameters than the number of trainable parameters. 
The cost increase due to raising the LoRA rank for a given model size is significantly smaller than the cost increase by enlarging model sizes. 
For example, increasing the LoRA rank from 8 to 64 nearly matches the performance as LoRA fine-tuning a 65B model with same rank, but only requires 50\% of 65B model's training cost. In practice we find that tuning 33B model provide a good trade-off between cost and performance.
% With a small fixed budget, we suggest that we use full-model fine-tuning to get a smaller model rather than using LoRA training to obtain a larger model, not only because of the much lower cost in the training stage, but also because of the high throughput in the serving stage. However, to push the SoTA performance with sufficient budget, larger models are necessary.

%In Table~\ref{tab:tuning_methods}(b), we compare different LMM training strategies. 
%It turns out that 

Different LoRA variations have similar performance, and 
QLoRA requires lower GPU memory cost and running-time cost than LoRA. When large models (\eg 65B) are trained with DeepSpeed ZeRO2 mode, they can fit into GPU with QLoRA, while yield the OOM issue with LoRA. In the experiments, we find that the hyperparameters of LoRA have a large impact of performance:$(i)$ Large learning rate and alpha value of LoRA improves the results significantly. For example, With the same rank=64, we reduce the learning rate=$2\times10^{-5}$ and alpha=16, the performance decrease from 71.8 to 65.5 on LLaVA-Bench. $(ii)$ Under the same setting, large ranks leads to little improvement. \eg we increase the rank from 64 to 128 and 512, it improves from 65.5 to 66.1 and 68.1, respectively. 

% \chunyl{empirical evidence/numbers are needed}

% as indicated by the numbers of trainable model parameters

% This observation suggests that conventional parameter-efficient fine-tuning techniques may not be suitable for adapting to multimodal models \cite{lora}. We will leave the exploration of further experiments as future work. 

% \yl{how about re-phrase it as "We have observed that the performance improves as we set a larger LoRA rank, This observation suggests that conventional parameter-efficient fine-tuning techniques may not be suitable for adapting to multimodal models \cite{lora}. We will leave the exploration of further experiments as future work. "}

% We also train 33B with larger LoRA rank (128 and 512) with 1 epoch, but found it still presents a gap compared to full fine-tuning on llava bench. Meanwhile, we observe that LoRA training generally takes longer time to converge. And performance gap with full finetuning is potentially due to the larger LoRA rank requires longer training time, which we do not have budget to experiment more.

\begin{table}[t!]
    \centering
    \begin{subtable}{1.0\textwidth} 
    \centering 
    \resizebox{0.99\textwidth}{!}{%
    \begin{tabular}{l|r|rr|rrrr|rr}
        \toprule
         & 7B & \multicolumn{2}{c|}{13B} & \multicolumn{4}{c|}{33B} &  \multicolumn{2}{c}{65B}\\
         LoRA Rank& Full  & 64 & Full & 8  & 64-QLoRA & 64  & Full & 64 & Full\\
        \midrule
        Performance $\uparrow$ & 65.9 & 70.1 & 70.1 & 70.3 & 71.6  & 71.8  & 72.0 & 72.2  &  72.3 \\
        Time (GPU Hours per node) $\downarrow$ & 1.3 & 2.1 & 2.3 & 4.62& 4.68  & 4.79   &  5.80   & 9.17  & 13.50  \\
        \# Trainable Parameters (B) $\downarrow$ & 7 & 0.26 & 13  & 0.06 & 0.49  & 0.49 & 33 & 0.81 &  65 \\
        \bottomrule
    \end{tabular}
    }
    % \caption{Comparisons of performance-cost trade-off on LLaVA-80K data. }
    \end{subtable}

    \vspace{1mm}
    \caption{The trade-off between performance and compute cost among different model sizes and traing methods on LLaVA-80K data. ``Full'' indicates the full-model fine-tuning. ``Time'' is reported as the total GPU hours to finish 1 epoch training (running time $\times$ \#GPUs) divided by 8 (\#GPUs per node).
    All models are trained on LLaVA-80K data, results are obtained through averaging 3 repeated evaluation runs with same set up on LLaVA-Bench. }
    %  \chunyl{correct the time row in this table}
    % \chunyl{What does "unfreeze embed" mean? I guess image encoder tuning? but please clarify}
    % \chunyl{Please fill in the number on LoRA as well.}}
    \label{tab:tuning_methods}
\end{table}

\paragraph{\textcircled{\raisebox{-0.9pt}{3}} A LMM with strong capabilities in both language and multimodal?}
We expand our evaluation in two aspects: $(i)$ MM-VET is added to measure the integrated multimodal capabilities of LMM; $(ii)$ The pure language ability of LMM is measured using Vicuna-80~\citep{vicuna} and MMLU~\citep{hendrycks2020measuring}, where the former evaluates the instruct-following ability in real-world language tasks, the latter evaluates the multilingual multi-task language ability. The results are shown in Table \ref{tab:eval_multimodal_and_langauge_tasks}, where all models are full-model fine-tuned. 

Compared to Vicuna which initializes the LLM weights of LLaVA, it is surprising to observe that LLaVA, after being trained solely on multimodal instruction data, exhibits a comparable language capability. Mixing language instruction data can boost LLaVA's multimodal ability, but not the language ability. %We find this result encouraging, and 
This is partially attributed to the inclusion of complex reasoning questions, and long-form answers in LLaVA-Instruct-158K, which helps maintain the language capabilities of LLaVA. 
We also train LLaVA-70B based on the LLaMA-2-70B-Chat checkpoint~\citep{touvron2023llama}, and find that mixed results on multimodal and language abilities. 
Interestingly, we improve LLaMA-2-70B-Chat by 2.4 points on MMLU, yielding an overall MMLU score of 65.1, which is the best performance for the 70B model size, according to ~\citep{wang2023far} and the Chatbot Arena Leaderboard~\footnote{\url{https://huggingface.co/spaces/lmsys/chatbot-arena-leaderboard}}. To the best of our knowledge, this is the first reported result which shows visual instructing tuning improve language ability of large-scale LMM.

% MTBench	MMLU
% Vicuna-13B	6.57	55.8
% LLaVA-13B	6.63	55.0

\begin{table}[t!]
    \centering
    \begin{tabular}{lc|cc|cc}
        \toprule
        \multirow{2}{*}{Model}  & \multirow{2}{*}{Data Mix}  & \multicolumn{2}{c|}{Multimodal}  & \multicolumn{2}{c}{Language} \\
          & & LLaVA-Bench & MM-VET & Vicuna-80  & MMLU \\
        \midrule
        Vicuna-${\text{13B}}$ &  -  & - & - & 79.9  & 55.8  \\
        \shortname{}-${\text{13B}}$ & \xmark & 70.1 & 32.5 & 79.6  &  55.0 \\
        \midrule
        Vicuna-${\text{33B}}$ &  -  & - & - & 85.6 & 59.0  \\
        \shortname{}-${\text{33B}}$ & \xmark & 72.0 & 32.9 & 85.3 & 56.1  \\
        \shortname{}-${\text{33B}}$ & \cmark & 73.9 & 34.1 &  80.3  & 58.6 \\ \midrule
        Vicuna-${\text{65B}}$ &  - & - & - &  83.2 & 62.5 \\
        \shortname{}-${\text{65B}}$ & \xmark & 72.3 & 35.5 & 84.5   & 62.6  \\ 
        \shortname{}-${\text{65B}}$ & \cmark & 74.2 &36.4 &  82.6 & 62.2 \\ 
        \midrule
        LLaMA-2-70B-Chat & - & - & - & 84.7  & 63.1 \\
        \shortname{}-${\text{70B}}$ & \cmark & 69.8 & 35.4 &  81.3 & {\bf 65.1} \\ 
        \bottomrule
    \end{tabular}
    \vspace{0.1in}
    \caption{Performance on both multimodal and language capabilities.}
    \label{tab:eval_multimodal_and_langauge_tasks}
\end{table}

\section{Conclusions and Limitations}
We present an empirical study of scaling the language model size for LMM. Our main findings are:
$(i)$ Scaling LMM consistently enhances model performance, resulting in significant improvements in language capabilities, primarily due to the increased LLM model size. We leave it to future work how to scale the vision encoder to enhance the visual capabilities and improve model performance on vision recognition and understanding tasks.
$(ii)$ Parameter-efficient methods such as LoRA/QLoRA are viable solutions to finetune large-scale LLMs for a good performance-cost trade-off in some real-world settings with limited GPU memory. We observe that LoRA/QLoRA's performance are comparable to that of fine-tuning the full model, establishing their effectiveness through significant cost reduction in both model training and serving.
$(iii)$ Our study of training data curation reveals that properly selecting image resolutions and mixing multimodal-language data for model training can significantly improve the performance of the resultant LMM. We also show for the first time that visual instruction tuning can improve LMM's language capability.
%, including image resolution selection and multimodal-language data mixing, reveals that both factors can significantly affect the behavior of LMMs. 
%Appropriate mixing can lead to enhanced performance in language or multimodal tasks. In some cases, visual instruction tuning can even improve LMM's language capability.
%However, we acknowledge that our scaling experiments and findings are still preliminary due to the lack of data scaling.
Note that the training datasets used in this study is small. So, our findings are still preliminary.
In future work, we will experiment using much larger datasets to investigate in detail whether and how different methods of training data selection and mixing can improve the quality of much larger LMM.

\bibliography{egbib}

\begin{thebibliography}{10}

\bibitem{alayrac2022flamingo}
Jean-Baptiste Alayrac, Jeff Donahue, Pauline Luc, Antoine Miech, Iain Barr,
  Yana Hasson, Karel Lenc, Arthur Mensch, Katherine Millican, Malcolm Reynolds,
  et~al.
\newblock Flamingo: a visual language model for few-shot learning.
\newblock {\em Advances in Neural Information Processing Systems},
  35:23716--23736, 2022.

\bibitem{awadalla2023openflamingo}
Anas Awadalla, Irena Gao, Josh Gardner, Jack Hessel, Yusuf Hanafy, Wanrong Zhu,
  Kalyani Marathe, Yonatan Bitton, Samir Gadre, Shiori Sagawa, et~al.
\newblock Openflamingo: An open-source framework for training large
  autoregressive vision-language models.
\newblock {\em arXiv preprint arXiv:2308.01390}, 2023.

\bibitem{dai2023instructblip}
Wenliang Dai, Junnan Li, Dongxu Li, Anthony Meng~Huat Tiong, Junqi Zhao,
  Weisheng Wang, Boyang Li, Pascale Fung, and Steven Hoi.
\newblock Instructblip: Towards general-purpose vision-language models with
  instruction tuning, 2023.

\bibitem{dettmers2023qlora}
Tim Dettmers, Artidoro Pagnoni, Ari Holtzman, and Luke Zettlemoyer.
\newblock Qlora: Efficient finetuning of quantized llms.
\newblock {\em arXiv preprint arXiv:2305.14314}, 2023.

\bibitem{gao2023llama}
Peng Gao, Jiaming Han, Renrui Zhang, Ziyi Lin, Shijie Geng, Aojun Zhou, Wei
  Zhang, Pan Lu, Conghui He, Xiangyu Yue, et~al.
\newblock Llama-adapter v2: Parameter-efficient visual instruction model.
\newblock {\em arXiv preprint arXiv:2304.15010}, 2023.

\bibitem{hendrycks2020measuring}
Dan Hendrycks, Collin Burns, Steven Basart, Andy Zou, Mantas Mazeika, Dawn
  Song, and Jacob Steinhardt.
\newblock Measuring massive multitask language understanding.
\newblock {\em arXiv preprint arXiv:2009.03300}, 2020.

\bibitem{hu2021lora}
Edward~J Hu, Yelong Shen, Phillip Wallis, Zeyuan Allen-Zhu, Yuanzhi Li, Shean
  Wang, Lu~Wang, and Weizhu Chen.
\newblock Lora: Low-rank adaptation of large language models.
\newblock {\em arXiv preprint arXiv:2106.09685}, 2021.

\bibitem{li2023otter}
Bo~Li, Yuanhan Zhang, Liangyu Chen, Jinghao Wang, Jingkang Yang, and Ziwei Liu.
\newblock Otter: A multi-modal model with in-context instruction tuning.
\newblock {\em arXiv preprint arXiv:2305.03726}, 2023.

\bibitem{li2023large}
Chunyuan Li.
\newblock Large multimodal models: Notes on {CVPR} 2023 tutorial.
\newblock {\em arXiv preprint arXiv:2306.14895}, 2023.

\bibitem{li2023multimodal}
Chunyuan Li, Zhe Gan, Zhengyuan Yang, Jianwei Yang, Linjie Li, Lijuan Wang, and
  Jianfeng Gao.
\newblock Multimodal foundation models: From specialists to general-purpose
  assistants.
\newblock {\em arXiv preprint}, 2023.

\bibitem{li2023blip}
Junnan Li, Dongxu Li, Silvio Savarese, and Steven Hoi.
\newblock Blip-2: Bootstrapping language-image pre-training with frozen image
  encoders and large language models.
\newblock {\em arXiv preprint arXiv:2301.12597}, 2023.

\bibitem{llava}
Haotian Liu, Chunyuan Li, Qingyang Wu, and Yong~Jae Lee.
\newblock Visual instruction tuning, 2023.

\bibitem{liu2023mmbench}
Yuan Liu, Haodong Duan, Yuanhan Zhang, Bo~Li, Songyang Zhang, Wangbo Zhao, Yike
  Yuan, Jiaqi Wang, Conghui He, Ziwei Liu, et~al.
\newblock Mmbench: Is your multi-modal model an all-around player?
\newblock {\em arXiv preprint arXiv:2307.06281}, 2023.

\bibitem{gpt4}
OpenAI.
\newblock Gpt-4 technical report, 2023.

\bibitem{touvron2023llama}
Hugo Touvron, Louis Martin, Kevin Stone, Peter Albert, Amjad Almahairi, Yasmine
  Babaei, Nikolay Bashlykov, Soumya Batra, Prajjwal Bhargava, Shruti Bhosale,
  et~al.
\newblock Llama 2: Open foundation and fine-tuned chat models.
\newblock {\em arXiv preprint arXiv:2307.09288}, 2023.

\bibitem{vicuna}
Vicuna.
\newblock Vicuna: An open-source chatbot impressing gpt-4 with 90\%* chatgpt
  quality.
\newblock \url{https://vicuna.lmsys.org/}, 2023.

\bibitem{wang2023far}
Yizhong Wang, Hamish Ivison, Pradeep Dasigi, Jack Hessel, Tushar Khot,
  Khyathi~Raghavi Chandu, David Wadden, Kelsey MacMillan, Noah~A Smith,
  Iz~Beltagy, et~al.
\newblock How far can camels go? exploring the state of instruction tuning on
  open resources.
\newblock {\em arXiv preprint arXiv:2306.04751}, 2023.

\bibitem{yang2023mmreact}
Zhengyuan Yang, Linjie Li, Jianfeng Wang, Kevin Lin, Ehsan Azarnasab, Faisal
  Ahmed, Zicheng Liu, Ce~Liu, Michael Zeng, and Lijuan Wang.
\newblock Mm-react: Prompting chatgpt for multimodal reasoning and action,
  2023.

\bibitem{yu2023mm}
Weihao Yu, Zhengyuan Yang, Linjie Li, Jianfeng Wang, Kevin Lin, Zicheng Liu,
  Xinchao Wang, and Lijuan Wang.
\newblock Mm-vet: Evaluating large multimodal models for integrated
  capabilities.
\newblock {\em arXiv preprint arXiv:2308.02490}, 2023.

\bibitem{zhu2023minigpt}
Deyao Zhu, Jun Chen, Xiaoqian Shen, Xiang Li, and Mohamed Elhoseiny.
\newblock Minigpt-4: Enhancing vision-language understanding with advanced
  large language models.
\newblock {\em arXiv preprint arXiv:2304.10592}, 2023.

\end{thebibliography}
\bibliographystyle{plain}

% \clearpage
% \appendix
% \input{subtex/appendix}

%%%%%%%%%%%%%%%%%%%%%%%%%%%%%%%%%%%%%%%%%%%%%%%%%%%%%%%%%%%%

\end{document}